\documentclass[10pt,twocolumn,letterpaper]{article}

\usepackage{cvpr}
\usepackage{times}
\usepackage{epsfig}
\usepackage{graphicx}
\usepackage{amsmath}
\usepackage{amssymb}
\usepackage{svg}
\usepackage{caption}
\usepackage{tabularx}
\usepackage{fbox}
\usepackage[export]{adjustbox}
\usepackage{booktabs}
\usepackage{tcolorbox}
\usepackage{listings}
\usepackage{xcolor}
\usepackage{tikz}
\usetikzlibrary{calc,shadows.blur,positioning}
\usepackage{capt-of}
\tcbuselibrary{skins}
\usepackage{booktabs}
\usepackage[margin=1in]{geometry}
\usepackage{array}
\newcolumntype{L}[1]{>{\raggedright\arraybackslash}p{#1}}

\usepackage[breaklinks=true,bookmarks=false]{hyperref}

\cvprfinalcopy 

\def\cvprPaperID{****} 

\begin{document}

\title{Predicting Winning Captions for Weekly New Yorker Comics}

\author{Stanley Cao\\
Stanford University\\
Department of Computer Science\\
{\tt\small stancao@stanford.edu}
\and
Sonny Young\\
Stanford University\\
Department of Chemistry\\
{\tt\small ssyoung@stanford.edu}
}

\maketitle

\begin{abstract}
Image captioning using Vision Transformers (ViTs) represents a pivotal convergence of computer vision and natural language processing, offering the potential to enhance user experiences, improve accessibility, and provide textual representations of visual data. This paper explores the application of image captioning techniques to New Yorker cartoons, aiming to generate captions that emulate the wit and humor of winning entries in the New Yorker Cartoon Caption Contest. This task necessitates sophisticated visual and linguistic processing, along with an understanding of cultural nuances and humor. We propose several new baselines for using vision transformer encoder-decoder models to generate captions for the New Yorker cartoon caption contest.
\end{abstract}

\section{Introduction}
Image captioning using Vision Transformers (ViTs) is a pivotal area of research, merging the domains of computer vision and natural language processing. The ability to automatically generate accurate and contextually relevant descriptions for images comes with numerous benefits, including enhancing user experience in various digital platforms, increasing accessibility for visually impaired users, and providing textual representations of visual data.

In this paper, we explore the complexities of applying image captioning techniques to New Yorker cartoons, particularly focusing on generating captions that closely emulate the wit and humor of winning entries in the New Yorker Cartoon Caption Contest. This task requires not only a sophisticated combination of visual and linguistic processing, but also the ability to caption images in a completely orthogonal way: the winning caption rarely describes the cartoon, but rather \textit{presupposes} a deep understanding of the scene along with the concomitant cultural references.  By aiming to create models that can effectively mimic human-like humor in captioning, we demonstrate potential advancements in the development of AI systems that have more holistic multi-modal understanding.


\begin{figure*}[ht]
\centering
\begin{minipage}{0.25\textwidth}
    \includegraphics[width=\linewidth]{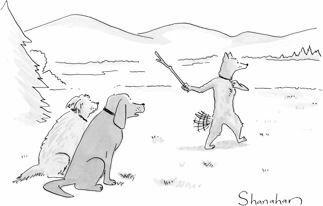}
\end{minipage}%
\hspace{0.1cm} 
\begin{minipage}{0.7\textwidth}
    \centering
    \begin{tabular}{@{}p{4.5cm} p{3.75cm} p{1.25cm} p{1.25cm}@{}}
        \toprule
        \textbf{Image Description} & \textbf{Uncanny Description} & \textbf{Entities} & \textbf{Caption} \\
        \midrule
        \small Three dogs are in a field in front of some rolling hills. One of the dogs is holding a stick and winding up, as if it's about to hit a baseball, while the other two dogs look on. & \small It's unusual that the dog is playing a sport like baseball because dogs don't have the cognitive capacity to play that human game. & \small Fetch, Dog, Anthropomorphism. & \small \textbf{He's his own best friend.} \\
        \bottomrule
    \end{tabular}
\end{minipage}
\caption{Detailed Information including selected metadata entries for New Yorker Cartoon Caption Contest \#102}
\label{dataset_example}
\end{figure*}

\section{Related Work}
Image captioning has so far been applied to numerous generative and classification problems dealing with  satellite data \cite{sun2024lightweight}, pathological diagnosis \cite{teramoto2024automated, qin2023slide}, or real-life object detection \cite{ahmed2023enhancing, xu2016show, lin2015microsoft} that primarily deal with natural images. Much success in these tasks has been attributed to attention models popularized by transformers \cite{vaswani2023attention} which embed sequential data in a parallel and expressive manner, and more recently, ViTs which are considered the state of the art architecture for image recognition \cite{dosovitskiy2021image}. 

The challenge is especially apparent when dealing with artificially constructed images, such as cartoons or comics. These images often contain abstract, stylized representations that may not correspond directly to real-world objects and may embed complex narratives and humor, which are particularly difficult for AI to parse and understand. Furthermore, ViTs are known to require significant pre-training data, close to 100 million images before its performance is comparable to a CNN \cite{dosovitskiy2021image}. Some have attempted to circumvent the large dataset issue by training ViTs without using natural data \cite{nakashima2021vision}. One particularly noteworthy model is LLaVA, a vision language model (VLM) that is capable of incorporating multi-modal information in its generated output \cite{Liu2023VisualIT}. \cite{liu2023improved} showcases that with simple modifications to the multi-modal projection layer and further finetuning on VQA datasets, the model is able to achieve improved performance on various downstream tasks. However,  evaluating the model beyond traditional question-answering tasks remains largely unexplored.

Natural Language Inference (NLI) is a key area that can enhance a model's ability to interpret language, which is crucial for improving how AI systems grasp concepts like humor \cite{chandrasekaran2016humor}, common sense \cite{commonsense7410649}, and overall sentence interpretation \cite{generalsentence6751319}. These developments suggest a path forward for refining AI approaches to better understand and engage with human communication nuances.

We focus on the New Yorker Cartoon Caption Contest, which requires humorous captioning that transcends pure information content.
Each week, participants are invited to submit captions for a cartoon, often characterized by its humorous, satirical, or ironic undertone, making the task of generating a fitting caption not just a test of understanding visual elements, but also of grasping subtle cultural and contextual nuances. Successfully modeling the New Yorker Cartoon captions would demonstrate a significant leap towards AI systems that can understand and generate human-like text responses. 

Notably, \cite{hessel2023androids} has explored this area of research, not only creating a dataset of New Yorker cartoons from prior contests, but proposing three different tasks to benchmark the capabilities revolving around multi-modal humor understanding: matching, quality ranking, and explanation. The authors report GPT-4 having superior performance in matching and quality ranking, achieving 84.5\% accuracy in matching a true caption among a pool of 5 fake captions. GPT-4 significantly outperformed all other methods, but it was unable to explain jokes as well as humans. The tasks proposed in \cite{hessel2023androids} are largely classification tasks, with the exception of their proposed explanation task. Since LLMs are unable to achieve human-level performance, these results suggest that there is room for further exploration in the generation of humorous captions.

\section{Dataset}
We utilized the New Yorker Caption Contest dataset available on Hugging Face \cite{newyorkernextmldataset, radev-etal-2016-humor, shahaf2015inside}. This dataset consists of cartoons that are used for three different tasks to assess the multi-modal capabilities of ViTs and encoder-decoder models. Since the task is to generate captions from the cartoon alone, much like what is expected of humans for the caption contest, we make use of the dataset that corresponds to the explanation task proposed in \cite{hessel2023androids}, which not only contains an cartoon and winning caption pair, but also important metadata such as human-annotated descriptions of the cartoon, an explanation for why the caption is humorous, important entities that are relevant to the cartoon, and leading questions that appear to motivate the winning caption. Figure \ref{dataset_example} shows a single data entry with selected metadata.

This dataset comprises approximately 2.6K data entries. We use the same train-validation-test splits as provided in \cite{newyorkernextmldataset}, with about 2.34k entries in the train set and about 130 entries in the validation and test set. We do not preprocess or normalize the cartoons in advance, as the ViTs used in this paper contain their own custom  image transformations / processing techniques (see Section \ref{method} for more details). As all cartoons are of different sizes, we have varying image resolutions, though most of ViTs presented in this paper normalize image resolution by padding the patches derived from the partitioning the image.

\section{Methodology and Experiments}
\label{method}
As our overarching goal is to generate the caption from the cartoon alone, we employ multi-modal vision encoder-decoder models for this task.

As one of the main findings in \cite{hessel2023androids}, computer vision serves as a bottleneck for top-quality explanation generation; models that have access to human-labeled cartoon scene descriptions perform significantly better. However, providing the model with both cartoon and human-annotated metadata does not accurately mimic the task of the New Yorker Cartoon Caption Contest --- the metadata for a given cartoon (e.g. image description) is inherently up for interpretation. It is the creative task of determining which aspects of the cartoon to satirically emphasize that is highlighted in such a contest. Thus, providing the model with metadata, which likely improves performance, means that we are leaking information to the model that a normal human participant in the contest does not have. In the remainder of this section, we outline 8 different training pipelines we devised for this task. We used the Hugging Face and Pytorch libraries for our tasks \cite{huggingface, Paszke2019PyTorchAI}

\subsection{CLIP + GPT2}
For our baseline, we implemented a CLIP-GPT2 encoder-decoder model that takes a cartoon as input and is required to output the caption. We employ the CLIP vision transformer model \cite{radford2021learning} to map all input cartoons into an multi-modal embedding space. From this embedding space, we use GPT2 \cite{Radford2019LanguageMA} as a decoder, translating this image embedding into a cartoon caption by using cross-attention between the encoder's output and the decoder's caption. In the experiments described in this section, we finetune our model end-to-end on the dataset for 10 epochs, using the AdamW optimizer with a learning rate of 5e-5. 
\label{clip_gpt2_base}

\subsubsection{Caption Only}
In this formulation, we ignore all metadata, simply tasking the model to generate the caption from the image alone. We include this approach as a baseline for this task.

\subsubsection{Caption + Metadata (all)}
In this formulation, we prepend the metadata to the target label, so that the caption is generated last. This training setup encourages the model to learn to describe the image and generate relevant metadata first. Due to the transformer architecture in decoder models, the model would use its attention mechanism to refer to previously self-generated metadata, culminating in a generated caption that is well-informed of the prior metadata. Figure \ref{fig:prepend_label} shows an example of synthesized target label.

\begin{figure}[ht]
    \begin{tabular}{@{} p{0.15\columnwidth} p{0.8\columnwidth} @{}}
    \textbf{Label}: & \texttt{scene: a field description: 3 dogs are out in a field. One is standing in front of the other two and preparing to throw a stick uncanny: It's unusual to see dogs holding items as they do not have oposable digits entities: Anthropomorphism, Dog, Baseball. caption: He's his own best friend.}
    \end{tabular}
    \caption{Modified target label with metadata prepended for the cartoon in Figure \ref{dataset_example}}
    \label{fig:prepend_label}
\end{figure}

\subsection{LLaVA-NeXT}
LLaVA-based models are variants of vision encoder-decoder models, which 
projects the input image onto an embedding space that has the same dimension as the language embedding space. By training this light-weight projection layer, LLaVA-NeXT is able to turn images into tokens, which are then subject to the same attention mechanisms as the textual data.

In \cite{Liu2023VisualIT}, they used a \textit{CLIP-ViT-L/14} to \textit{Vicuna} LLM encoder-decoder model. However, in our experiments, we use the LLaVA-NeXT model. In the original paper, the authors use a \textit{CLIP-ViT-L-336px} to \textit{Vicuna} LLM \cite{liu2024llavanext}, but we make use of a variant hosted on Hugging Face, which replaces \textit{Vicuna} with \textit{Mistral} LLM. 

\subsubsection{0-shot Setting}
We evaluate the efficacy of LLaVA-NeXT on this task in a 0-shot setting. Figure \ref{fig:llava_0shot} shows the 0-shot prompt we used to query the model, and we extract its response as the generated caption. The model is not presented any examples in advance, and must execute the task using only its prior knowledge base.

\begin{figure}[ht]
    \begin{tabular}{@{} p{0.15\columnwidth} p{0.8\columnwidth} @{}}
    \textbf{Input}: & \texttt{[INST] <image> Please write a winning caption for the provided cartoon in the New Yorker caption contest. [/INST]}
    \end{tabular}
    \caption{0-shot prompt for LLaVA-NeXT and GPT-4V. The \texttt{<image>} token refers to the placement of the image embedding for the LLaVA-NeXT Model.}
    \label{fig:llava_0shot}
\end{figure}

\subsubsection{5-shot Setting}
In addition to the 0-shot setting, we evaluate the efficacy of LLaVA-NeXT on this task in a 5-shot setting. Specifically, we show the model 5 examples, as demonstrated in Figure \ref{fig:llava_5shot}, with the intention to encourage the model to learn the style of the task.

\begin{figure}[ht]
    \small
    \begin{tabular}{@{} p{0.15\columnwidth} p{0.8\columnwidth} @{}}
    \textbf{Input}: & \texttt{[INST] <image1> [/INST] CAPTION 1} \\
    \textbf{Input}: & \texttt{[INST] <image2> [/INST] CAPTION 2} \\
    \textbf{Input}: & \texttt{[INST] <image3> [/INST] CAPTION 3} \\
    \textbf{Input}: & \texttt{[INST] <image4> [/INST] CAPTION 4} \\
    \textbf{Input}: & \texttt{[INST] <image5> [/INST] CAPTION 5} \\
    \textbf{Input}: & \texttt{[INST] <image> Please write a winning caption for the provided cartoon in the New Yorker caption contest.[/INST]}
    \end{tabular}
    \caption{5-shot prompt for LLaVA-NeXT. The \texttt{<image>} token refers to the placement of the image embeddings. The final user input asks the model to generate a caption after it has seen 5 previous examples of human-written winning captions.}
    \label{fig:llava_5shot}
\end{figure}

\subsubsection{Chain-of-Thought Prompting}
Chain-of-Thought (CoT) is a prompt engineering technique designed to enhance language models' performance on logic, calculation, and decision-making tasks by structuring the input prompt to simulate human reasoning \cite{CoTPrompting}. We provide LLaVA-NeXT with a series of intermediate reasoning steps that encourages more sophisticated decision-making. Figure \ref{fig:llava_gpt4_cot} showcases our proposed Chain-of-Thought prompting technique.

\begin{figure}[ht]
    \small
    \begin{tabular}{@{} p{0.1\columnwidth} p{0.8\columnwidth} @{}}
    \textbf{Input}: & 
    \texttt{[INST]
    <image>
    Please write an engaging, witty entry for the New Yorker Caption Contest based on this image. You are competing against 5,000 entrants from across the world, many of whom have made hundreds of previous submissions.} \\
    \vspace{0.5em} \\
    & \texttt{Please follow these rules:} \\
    \vspace{0.5em} \\
    & \texttt{1. First, describe the comic in as much objective detail as you can, subject to a limit of 4 sentences max. Leave no stone unturned and describe all main objects and characters.}
    \vspace{0.5em} \\
    & \texttt{2. Then, list as many uncanny aspects of the comic as you can think of (list at least 5). Then, for each one, explain why it is uncanny and why the comic artist likely inserted it. Think about what makes each aspect of the unusual occurrence humorous.} \\
    \vspace{0.5em} \\
    & \texttt{3. Based on your descriptions above and using great taste, come up with 5 potential caption concepts, each of which should be your best work and maximally diverse. Then, roast each one in 2-3 sentences.} \\
    \vspace{0.5em} \\
    & \texttt{4. Write 5 improved concepts after that, explaining why each is better.} \\
    & \texttt{[/INST]} \\
    \vspace{1em}\\
    \textbf{LLM:} & \texttt{GENERATED RESPONSE} \\
    \vspace{1em}\\
    \textbf{Input:} & \texttt{[INST] Thanks! Now, select the best caption from all the versions you've written above. Write only the caption as a single line, with no other text. [/INST]}
    \end{tabular}
    \caption{Chain-of-Thought prompting for both LLaVA-NeXT and GPT-4V.}
    \label{fig:llava_gpt4_cot}
\end{figure}

\subsubsection{Finetuning with QLoRA}
We finetuned LLaVA-NeXT on our training set in a question-answer format to remain consistent with the model's prior pre-training. Low-Rank Adaptation (LoRA) is a popular finetuning mechanism for LLMs that significantly reduces the number of trainable parameters without losing the benefit of downstream finetuning \cite{hu2022lora}. LoRA approximates updates to weight matrices with a low-rank matrix decomposition.
Thus, for a given weight matrix $W_0 \in \mathbb{R}^{d \times k}$, LoRA constrains the update $\Delta W$ to be approximated by a low-rank matrix decomposition $\Delta W \approx BA$, where $B \in \mathbb{R}^{d \times r}$ and $A \in \mathbb{R}^{r \times k}$, and $r \ll \min(d, k)$. This means that the update to weight matrices becomes 
\[
h = (W_0 + \Delta W)x \approx (W_0 + BA)x.
\]
During training, $W_0$ is frozen and does not receive gradient updates, while A and B are trainable
parameters. 

Due to compute limitations, we use the 7B parameter model as opposed to the 34B parameter model. In addition, we implemented quantized LoRA (QLoRA) and LoftQ initialization \cite{dettmers2023qlora, li2024loftq} to further optimize memory usage. We finetuned the LLaVA-NeXT model end-to-end on an A100 80GB GPU for 10 epochs with an $r$ value of 32, learning rate of 1e-4, $\alpha$ of 16 on all linear layers.

\subsection{GPT-4V}

We repeated the same evaluation of this task using the GPT-4V multi-modal model with both 5-shot and complex CoT mechanisms. We used the OpenAI API to query GPT-4V and utilized the same prompts as our evaluation for LLaVA-NeXT.


\begin{figure*}[htbp]
    \centering
    \begin{tikzpicture}
        \node[anchor=center] (image) at (0, 0.5) {
                \centering
                \includegraphics[width=5cm]{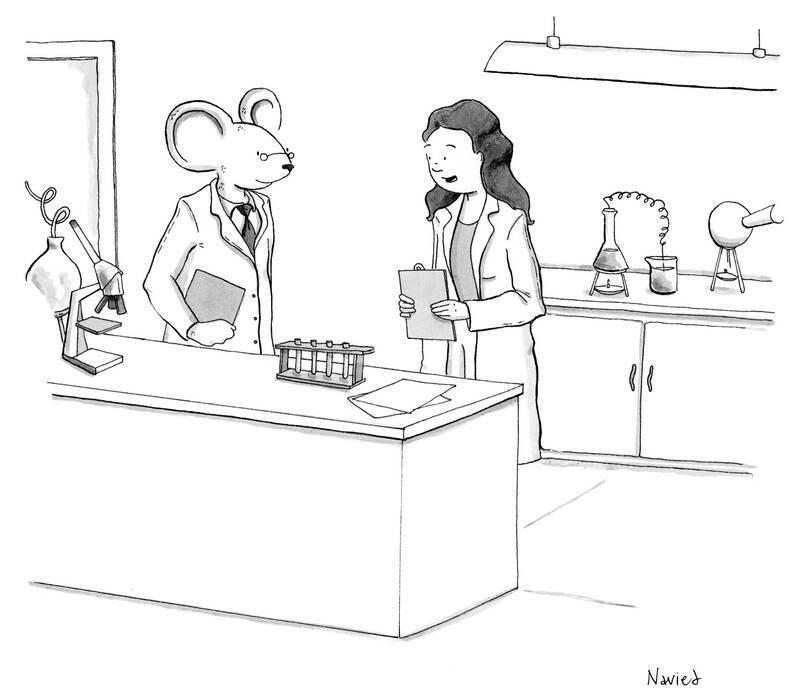}
        };
        
        \definecolor{boxcolor}{HTML}{FFFFFFF}

        \node[anchor=south] (box1) at (0, 3) {\begin{tcolorbox}[colback=boxcolor, boxrule=0.5pt, width=5.25cm, height=2.5cm]\textbf{GPT-4V (5-shot)}\newline I'm here for the job interview. I have extensive experience running through mazes.
        \end{tcolorbox}};
        \node[anchor=south] (box2) at (5.5, 3) {\begin{tcolorbox}[colback=boxcolor, boxrule=0.5pt, width=5.25cm, height=2.5cm]\textbf{LLaVA-NeXT (0-shot)}\newline A moment of scientific discovery, where the mouse's curiosity meets the scientist's expertise.
        \end{tcolorbox}};
        \node[anchor=south] (box3) at (-5.5, 3) {\begin{tcolorbox}[colback=boxcolor, boxrule=0.5pt, width=5.25cm, height=2.5cm]\textbf{GPT-4V (complex CoT)}\newline Should I be concerned that my new pharmacist is more interested in running trials than running prescriptions?\end{tcolorbox}};
        \node[anchor=south] (box4) at (5.5, -1.6) {\begin{tcolorbox}[colback=boxcolor, boxrule=0.5pt, width=5.25cm, height=4cm]\textbf{LLaVA-NeXT (5-shot)}\newline As the lab assistant, I always find it strange that the mouse holds the book... but never reads it.\end{tcolorbox}};
        \node[anchor=south] (box5) at (-5.5, -1.6) {\begin{tcolorbox}[colback=boxcolor, boxrule=0.5pt, width=5.25cm, height=4cm]\textbf{LLaVA-NeXT Finetuned (0-shot)}\newline A symposium on advanced hearing aids, featuring the latest in supersonic frequency bending technology.\end{tcolorbox}};
        \node[anchor=north] (box6) at (0, -2) {\begin{tcolorbox}[colback=boxcolor, boxrule=0.5pt, width=5.25cm, height=2.5cm]\textbf{CLIP-GPT2 (caption only)}\newline This is the first time I've had a neck injury.\end{tcolorbox}};
        \node[anchor=north] (box7) at (5.5, -2) {\begin{tcolorbox}[colback=boxcolor, boxrule=0.5pt, width=5.25cm, height=2.5cm]\textbf{CLIP-GPT2 (all metadata)}\newline Can't you just let him finish?\end{tcolorbox}};
        \node[anchor=north] (box8) at (-5.5, -2) {\begin{tcolorbox}[colback=boxcolor, boxrule=0.5pt, width=5.25cm, height=2.5cm]\textbf{LLaVA-NeXT (complex CoT)}\newline Who knew mice had such a high IQ?\end{tcolorbox}};
    \end{tikzpicture}
    \centering
    \captionsetup{justification=centering}
    \caption{Generated Captions for New Yorker Caption Contest \#54. \newline \textbf{Winning caption}:  
    \textit{And, when you get hungry, the cafeteria is to your right, left, left, right, left, straight, right, straight, left, and then you push on the big lever.}}
    \label{fig:3x3_layout}        

\end{figure*}

\section{Results}
We compare our generated captions to the human-written captions via automated evaluation metrics (e.g., BLEU and ROUGE) and manual quality examination. 

\subsection{BLEU and ROUGE Scores}
A well-known metric for automated quality evaluation of machine generated text is the Bilingual Evaluation Understudy (BLEU) score metric, which is designed to evaluate the quality of text from one natural language to another\cite{10.3115/1073083.1073135, lin-och-2004-orange}. All models performed poorly with respect to the BLEU score metric. Recall-Oriented Understudy for Gisting Evaluation (ROUGE) is another popular scoring metric for evaluating automatic summarizing and machine translation. Figure \ref{tab:rouge_metrics} shows the ROUGE scores for each of our models on the test set. While our models performed poorly with respect to these automated metrics, we will later discuss why these automated metrics, which are designed for machine translation / summarization tasks, do not serve as a good metric for this usecase.

\subsection{Manual Quality Evaluation}

To surface a more accurate metric for this task, we propose a criteria called \textit{SS-SCORE} metric. For each of the 130 test images, we compare all the machine-generated captions with each other, using the human-written caption as a reference, and select the best machine-generated caption. The quality of the generated caption were based on the following criteria:

\begin{enumerate}
\item Captions should contain descriptive information about the cartoon that extends beyond listing features in the image. An observer who examines the image then subsequently internalizes the caption should most often understand the meaning behind why a situation was described the way it was or why a character would have contributed their specific dialogue. 
\item Good captions do not necessarily have to exhaustively describe the situation, but rather convey humor, wit, or a voice that aptly fits within the context of the scene.
\item A caption that addresses similar themes as the winning caption in the associated contest may lead to a more favorable selection criteria on the grounds of resembling an officially winning entry.
\end{enumerate}

After tallying each test example, for each model, we aggregate the statistics by calculating the number of times that model generated a caption that was deemed the best, normalizing by the total number of test examples to obtain a proportion which we equate to the \textit{SS-SCORE}. These metrics can be found in Figure \ref{tab:rouge_metrics}.



\section{Discussion}
\subsection{Automated Metrics}
From Figure \ref{tab:rouge_metrics}, we note that according to the various types of ROUGE scores, the CLIP-GPT2 model performs best, and by prepending the metadata to the target labels, we achieve a marginally higher ROUGE score. However, automated evaluation metrics such as BLEU and ROUGE do not accurately capture the quality of machine-generated captions. Since BLEU and ROUGE both operate on $n$-grams, these automated metrics are capturing the commonalities between machine generated and human-written captions at the individual word level without regards to semantic meaning. For a task as sophisticated as this one, word-level similarity is highly uncorrelated with caption quality because there are multiple ways to write humorous captions. Rather than regurgitating the content of the cartoon, contest participants are required to execute complex reasoning that relies on social and cultural nuances related to the cartoon, and based on the diverse range of high quality submissions for a single image, there is no single method for generating a good caption.

\subsection{Manual Quality Evaluation}
Based on \textit{SS-SCORE}, we notice that GPT-4V in the 5-shot setting performs the best, with GPT-4V in the Chain-of-Thought setting performing second best. This suggests that larger parameter models are able to perform better on this task, likely because they are trained on more data and have a larger knowledge base. GPT-4 models are constantly finetuned on the most recent data, which makes them particularly knowledgeable about both historic and current events; this knowledge is particularly valuable for this cartoon captioning contest, since cartoons from The New Yorker are often timely caricatures of socio-cultural situations. 

It is particularly interesting that the 5-shot setting performs almost twice as well as the Chain-of-Thought setting. This suggests that GPT-4V is capable of few-shot learning, and presenting examples of cartoon-caption pairs to GPT-4V appears more helpful for the model to adapt to the specific tone typically found in high quality cartoon captions compared to Chain-of-Thought prompting. Additionally, GPT-4V demonstrated other unique ways of expressing creative generated captions such as the use of puns and double entendres (e.g. a ``full-blown service'' for a tuba-playing waiter) or references to movies or literature (e.g. ``Cloudy with a Chance of Armageddon'' for a cloud with a man's face referring to the movie \textit{Cloudy with a Chance of Meatballs}).

\subsection{Qualitative Evaluation}
Figure \ref{fig:3x3_layout} shows one particular test example with the generated captions from all of our models. We also showcase the winning caption selected by The New Yorker magazine.

We note that although the CLIP-GPT2 models are able to mimic the correct tone and style of high quality captions, they are unable to capture the semantic content of the image. This is likely because the CLIP vision model was trained on natural images, and these cartoons do not exhibit the same color distribution as those natural images. We found that even with fine-tuning, CLIP is unable to capture the proper image features for GPT2 to decode. Moreover, the CLIP-GPT2 encoder-decoder model is far from the state-of-the-art vision transformer models, and due to the relatively small size of the model, it is highly unlikely that it will have the vast social knowledge to create humorous captions.

We also note that LLaVA-NeXT in the 0-shot setting has a tendency to create a \textit{title} for the cartoon rather than a caption for it. Even when we finetune LLaVA-NeXT on the dataset, the model still has the tendency to generate a title. We hypothesize that this is because LLaVA-NeXT does not understand the prompt, and it has no prior knowledge of The New Yorker cartoon caption contest. Thus, it is likely that ``New Yorker'' was tokenized into an unknown token which has zero knowledge about the magazine and its history, or at the very most has only surface-level knowledge of that linguistic entity. 

However, in the 5-shot setting LLaVA-NeXT is able to generate a caption that more closely mimics what could be found in a high quality caption. It begins to use pronouns like in the sentence ``I always find it strange...'' which indicates that it is learning to understand that high quality captions are typically statements made by characters in the cartoon. Moreover, the Chain-of-Thought setting shows further improvements, with the generated caption demonstrating a clear understanding that character dialogue is important for humorous captions.

Finally, the GPT-4V models perform the best. As discussed before, GPT-4V models have the most recent and largest knowledge database, and are able to successfully incorporate that knowledge into the caption. This is exhibited in its generated captions in Figure \ref{fig:3x3_layout}, which not only shows its understanding of the style of the caption, but also an understanding of what mice typically do in laboratory settings. Thus, we believe that GPT-4V is the current state-of-the-art model for this cartoon captioning task.

\begin{figure*}[h]
    \centering
    \begin{tabular}{lcccc}
      \toprule
        & ROUGE-1 & ROUGE-2 & ROUGE-L & SS-SCORE \\
      \midrule
      GPT-4V (CoT) & 0.0653 & 0.0049 & 0.0591 & 0.19 \\
      GPT-4V (5-shot) & 0.0648 & 0.0019 & 0.0598 & \textbf{0.37} \\
      LLaVA-NeXT (0-shot) & 0.0609 & 0.0006 & 0.0568 & 0.09 \\
      LLaVA-NeXT Finetuned (0-shot) & 0.0637 & 0.0036 & 0.0575 & 0.05 \\
      LLaVA-NeXT (5-shot) & 0.0586 & 0.0024 & 0.0540 & 0.01 \\
      LLaVA-NeXT (CoT) & 0.0672 & 0.0022 & 0.0617 & 0.04 \\
      CLIP-GPT2 (caption only) & 0.0733 & \textbf{0.0062} & 0.0665 & 0.11 \\
      CLIP-GPT2 (all metadata) & \textbf{0.0735} & 0.0039 & \textbf{0.0673} & 0.14 \\
      \bottomrule
    \end{tabular}
    \caption{Evaluation metrics for all models}
    \label{tab:rouge_metrics}
\end{figure*}


\section{Future Steps}
\label{sec:future_work}

The task of generating humorous and contextually relevant captions for New Yorker cartoons presents unique challenges that require sophisticated AI models. While our current work demonstrates significant advancements, several avenues for future exploration could enhance the capabilities and performance of such models. Our experiments with models like GPT-4V and LLaVA-NeXT highlight the importance of model size and the breadth of pre-training data. Future work could involve training even larger models, which inherently possess a more extensive knowledge base and greater capacity for understanding nuanced contexts. Utilizing models like the forthcoming GPT-5 could potentially offer improvements in generating high-quality, humorous captions. Additionally, leveraging distributed training on multiple GPUs or TPUs can facilitate the training of these larger models more efficiently.

Prompt engineering plays a crucial role in guiding the model to produce desirable outputs. Our use of few-shot learning and Chain-of-Thought (CoT) prompting has shown promising results, but further refinement and innovation in prompt design could yield even better performance. Especially for the LLaVA-NeXT model, we noticed many responses misinterpreted the task at hand by describing the image like a traditional caption more often than generating a line that has character or wit. Another option to make the captions more humorous and relevant is to explore hierarchical prompting, where prompts are dynamically adjusted based on intermediate model responses.

While the New Yorker Caption Contest dataset provides a robust foundation, curating additional datasets specifically aimed at caption generation could offer substantial benefits. These datasets could include a broader range of cartoons and comics, along with detailed annotations on humor, context, and cultural references. Collaborating with cartoonists and humorists to annotate datasets with insights into why certain captions are humorous could provide invaluable training data for models.


{\small
\bibliographystyle{ieee}
\bibliography{egbib}
}

\end{document}